%% file: icra2017.tex
\documentclass[letterpaper,10pt,conference]{ieeeconf}
\IEEEoverridecommandlockouts
\usepackage[utf8]{inputenc}
\usepackage{algorithm,algorithmic,amsmath,amssymb,caption,color,graphics,
            graphicx,hyperref,multirow,subfig,times,url,wrapfig}
\usepackage[binary-units]{siunitx}
\usepackage{balance}
\usepackage[table,xcdraw]{xcolor}
\include{defs}

\newcommand{\SB}{SUPERball}
\title{\LARGE \bf
    Deep Reinforcement Learning for Tensegrity Robot Locomotion
}

\author{Marvin Zhang$^{1,*}$, Xinyang Geng$^{1,*}$, Jonathan Bruce$^{2,*}$,
    Ken Caluwaerts$^{3}$, Massimo Vespignani$^{4}$,\\%
    Vytas SunSpiral$^{4}$, Pieter Abbeel$^{1,5,6}$, Sergey Levine$^{1}$%
    \thanks{$^{*}$These authors contributed equally to this work.}%
    \thanks{$^{1}$Department of Electrical Engineering and Computer Sciences,}%
    \thanks{\hspace{4px}University of California, Berkeley, Berkeley, CA 94720}%
    \thanks{$^{2}$Department of Computer Engineering,}%
    \thanks{\hspace{4px}University of California, Santa Cruz, CA 95064}%
    \thanks{$^{3}$Autodesk, Inc., San Francisco, CA 94111}%
    \thanks{$^{4}$Nasa Ames Research Center, Moffett Field, CA 94035}%
    \thanks{$^{5}$OpenAI, San Francisco, CA 94110}%
    \thanks{$^{6}$International Computer Science Institute, Berkeley, CA 94704}%
}

\begin{document}

\maketitle
\thispagestyle{empty}
\pagestyle{empty}

\begin{abstract}

Tensegrity robots, composed of rigid rods connected by elastic cables, have a
number of unique properties that make them appealing for use as planetary
exploration rovers. However, control of tensegrity robots remains a difficult
problem due to their unusual structures and complex dynamics. In this work, we
show how locomotion gaits can be learned automatically using a novel extension
of mirror descent guided policy search (MDGPS) applied to periodic locomotion
movements, and we demonstrate the effectiveness of our approach on tensegrity
robot locomotion. We evaluate our method with real-world and simulated
experiments on the \SB{} tensegrity robot, showing that the learned policies
generalize to changes in system parameters, unreliable sensor measurements, and
variation in environmental conditions, including varied terrains and a range of
different gravities. Our experiments demonstrate that our method not only learns
fast, power-efficient feedback policies for rolling gaits, but that these
policies can succeed with only the limited onboard sensing provided by \SB{}'s
accelerometers. We compare the learned feedback policies to learned open-loop
policies and hand-engineered controllers, and demonstrate that the learned
policy enables the first continuous, reliable locomotion gait for the real \SB{}
robot. Our code and other supplementary materials are available from
\url{http://rll.berkeley.edu/drl\_tensegrity}

\end{abstract}

\section{Introduction}
\label{sec:intro}

Tensegrity robots are a class of robots that are composed of rigid rods
connected through a network of elastic cables. These robots are lightweight, low
cost, and capable of withstanding significant impacts by deforming and
distributing force across the entire structure. These properties make them a
promising option for future planetary exploration missions, as their compliance
protects the robot and its payload during high-speed descent and landing, and
may also allow for greater mobility and robustness during exploration of rugged
and dangerous environments~\cite{bruce2014design}.

\begin{figure}
    \centering
    \setlength{\unitlength}{0.5\columnwidth}
    \includegraphics[width=\linewidth]{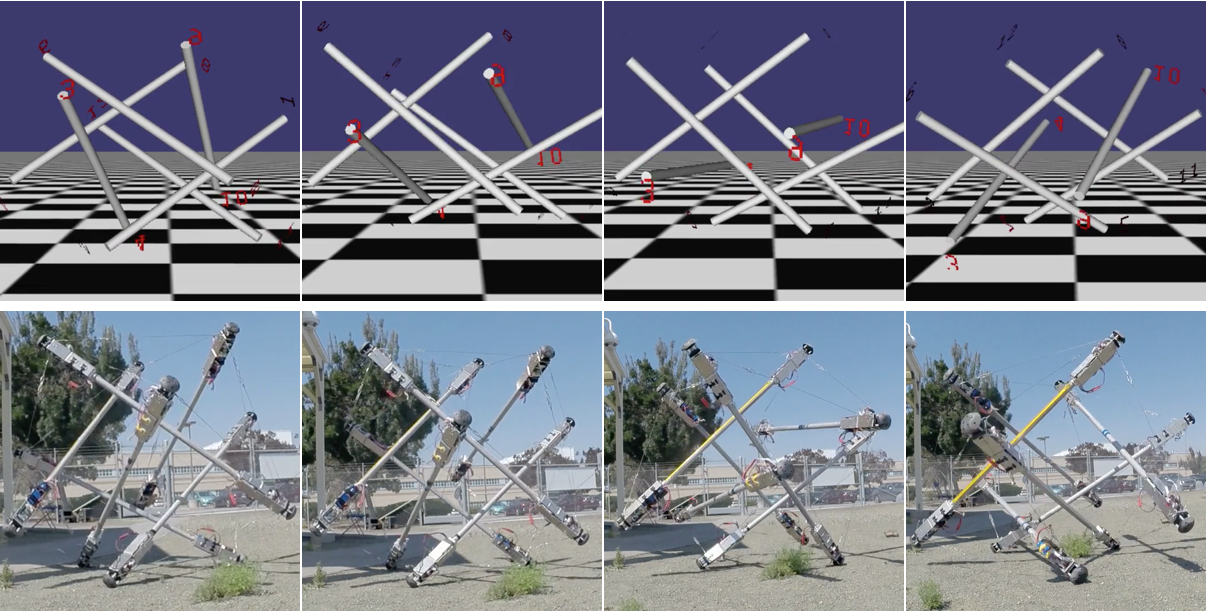}
    \caption{
        \label{fig:rolling}
        Left to right: a rolling motion toward the camera performed by the \SB{}
        simulation (top) and the physical \SB{} robot (bottom). The rolling gait
        is learned from scratch in simulation with our proposed algorithm, and
        uses only the onboard accelerometer sensors for feedback.
    }
\end{figure}

However, efficient locomotion for tensegrity robots is a challenging problem.
Such robots are typically controlled through actuation of motors that extend and
contract their cables, thereby changing their overall shape. Many such systems
are underactuated because there are usually more cables than motors.
Furthermore, actuating one motor can change the entire shape of the robot,
leading to complex, highly coupled dynamics. As such, hand-engineering
locomotion controllers is unintuitive and time-consuming and, as our
experimental results in Section~\ref{sec:results} show, such controllers often
do not generalize well to different environments. In particular, since the goal
of these robots involves deployment to celestial bodies with vastly different
terrains, gravities, and compositions, it is preferable to have policies be
automatically generated for each environment rather than to independently
hand-engineer controllers for each setting. This motivates using learning
algorithms to automatically discover successful and efficient behavior.

Robotic learning methods have previously produced successful policies for tasks
such as locomotion for bipeds and
quadrupeds~\cite{tzs-spgrl-04,kp-pgrlf-04,cspd-bolg-15,kn-ldquad-11,kbps-temp-09,mmea-darwin-16}.
These methods, however, typically require hand-engineered policy classes, such
as a linear function approximator using a set of hand-designed features as
input~\cite{tzs-spgrl-04}.  For many tensegrity systems, it is difficult to
design suitable policy classes, since the structure of a successful locomotion
strategy might be highly complex. We illustrate this in
Section~\ref{sec:results}, by demonstrating that it is desirable to have a
representation that is closed-loop, since open-loop control, though simpler to
design and implement, does not generalize as well to changes in terrain,
gravity, and other environmental and robot parameters.

Some more recent methods learn deep neural network policies that are successful
for tasks such as grasping with robotic arms and bipedal
locomotion~\cite{slmja-trpo-15,lhphe-ccdrl-16}, and such policies are more
expressive and require less hand-engineering compared to policy classes used in
previous methods. One such method, which we extend in this work, is mirror
descent guided policy search (MDGPS), a recently developed algorithm that frames
the guided policy search (GPS) alternating optimization framework as approximate
mirror descent~\cite{ml-gpsam-16}. We choose MDGPS in our work because it allows
us to learn deep neural network policies while maintaining sample efficiency,
and it presents a natural extension to periodic locomotion tasks which we
describe in Section~\ref{sec:chains}.

A key problem for locomotion tasks is the difficulty of establishing stable
periodic gaits, and this is exacerbated for tensegrity robots due to their
complex dynamics and unusual control mechanisms. As shown in our experiments,
near-stable behavior with even small inaccuracies can lead to compounding errors
over time, and will not be successful in producing a continuous periodic gait.
Previous algorithms have dealt with this problem by establishing periodicity
directly through the choice of policy
class~\cite{kp-pgrlf-04,gpw-fbwrc-06,emmnc-cpg-08}, utilizing a large number of
samples~\cite{slmja-trpo-15}, or initializing from
demonstrations~\cite{lk-gps-13,la-lnnpg-14}. Instead, we handle this challenge
by sequentially training several simple policies that demonstrate good behavior
from a wide range of states, and then learning a policy that reproduces the gait
of all of the sequential policies for a successful periodic behavior. The
resulting algorithm learns a policy from scratch for locomotion of a tensegrity
robot, a task that exhibits periodicity over long time horizons. We demonstrate
through our experiments that this learned policy is capable of efficient,
continuous locomotion in a range of different conditions by learning appropriate
feedbacks from the robot's onboard sensors.

The main contribution of this paper is a method for automatically learning
locomotion policies represented by general-purpose neural networks, which we
demonstrate by learning a gait for the Spherical Underactuated Planetary
Exploration Robot ball (\SB{}), the tensegrity robot shown rolling in
Figure~\ref{fig:rolling}. To this end, we extend the MDGPS algorithm so as to
make it suitable for learning long, periodic gaits, by training groups of
sequential policies as supervision for learning a successful neural network
policy. Our experimental results show that our method learns efficient rolling
behavior for \SB{} both in simulation and on the real physical robot. We make
comparisons between the learned policies and two open-loop representations, one
learned and one hand-engineered, to demonstrate the benefits of learning and
feedback for fast and reliable locomotion.

\section{Related Work}
\label{sec:related}

Early work in tensegrity research was focused on modeling the statics of
tensegrity structures~\cite{2003Tensegrity:-Str,Juan2008,Arsenault:2008bh}. 
This led to the development of kinematic controllers which enable quasi-static
locomotion~\cite{kimrobust}. More recent work has developed dynamic locomotion
controllers~\cite{graellsrovira2009} in simulation using the NASA Tensegrity
Robotics Toolkit (NTRT)~\cite{sunspiralsoftware}. Iscen~\etal used
coevolutionary learning that exploited the symmetry of a simulated \SB{}-like
robot to learn an efficient rolling controller~\cite{iscen2014flop}. This
controller required 24 actuators on the robot, which does not match the current
robot's 12 actuators. We also do not exploit the symmetry of \SB{} as this
strategy is less reliable on the real robot. A snake-like tensegrity robot
learned different locomotion gaits which utilized Central Pattern
Generators~\cite{mirletzsoftrobotics}, which were then managed by a neural
network to achieve goal directed behavior over various terrains. However, they
utilized Monte Carlo simulation techniques requiring thousands of trials to
learn their policies. This makes it impractical to learn on real hardware.
Mirletz~\etal successfully transferred a learned policy to a real robotic
prototype~\cite{mirletz2015towards}, but the policy parameters required
hand-tuning.

Previous work in robotic locomotion has produced successful bipedal locomotion
using passive-dynamic walkers~\cite{crtw-pdw-05} and virtual model
control~\cite{pctdp-vmc-01}, as well as spring-mass running based on biological
models~\cite{sgh-slr-03}. However, these works used analytic models built
through human insight and simplified models, and are therefore difficult to
generalize to radically different systems such as tensegrities. Robotic learning
methods have successfully learned locomotion policies for
bipeds~\cite{tzs-spgrl-04,gpw-fbwrc-06,emmnc-cpg-08} and
quadrupeds~\cite{kp-pgrlf-04}. These methods require careful consideration and
design of the policy class, which typically have fewer than 100
parameters~\cite{dnp-spsr-13}. More recent work in deep reinforcement learning
has learned deep neural network policies for bipedal
running~\cite{slmja-trpo-15} and 2D systems such as swimmers and
hoppers~\cite{lk-gps-13,la-lnnpg-14}.  These methods, however, typically require
a large number of samples corresponding to weeks of training
time~\cite{slmja-trpo-15}, or initialize from
demonstrations~\cite{lk-gps-13,la-lnnpg-14}.

Several methods have been proposed for the direct transfer of policies learned
in simulation to the real world. Cutler~\etal used a transition from simple to
complex simulations, which are assumed to increase in both cost and fidelity to
the real-world setting, to reduce the amount of training time needed in the real
world~\cite{cwh-sim-16}. Mordatch~\etal optimized trajectories through an
ensemble of models perturbed with noise to find trajectories that could be
successfully transferred to the real world setting~\cite{mlt-encio-15}. Our
approach also introduces noise into the simulation, and we found that this can
help in training policies that are successful in a wide range of simulated
terrain and gravity settings.

\section{Background}
\label{sec:background}

In this section, we present background on tensegrity robots, as well as policy
search methods. In particular, we describe the MDGPS
algorithm~\cite{ml-gpsam-16}, which we extend in this work to optimize periodic
locomotion gaits.

\subsection{Tensegrity Robotics}
\label{sec:tensegrity}

Tensegrity (\emph{tensile-integrity}) structures are free-standing structures
with axially loaded compression elements in a network of tension elements.
Ideally, each element of the structure experiences either pure axial compression
or pure tension~\cite{BuckminsterFuller1975,Snelson1965}. The absence of bending
or shear forces allows for highly efficient use of materials, resulting in
lightweight yet robust systems. Because the rods are not directly connected,
tensegrities have the unique property that externally applied forces distribute
through the structure via multiple load paths. This property creates a soft
robot out of inherently rigid materials. Since there are no rigid connections
within the structure, there are also no lever arms to magnify forces. The result
is a global level of robustness and tolerance to forces applied from any
direction.

\begin{figure}
    \centering
    \setlength{\unitlength}{0.5\columnwidth}
    \includegraphics[width=0.9\linewidth]{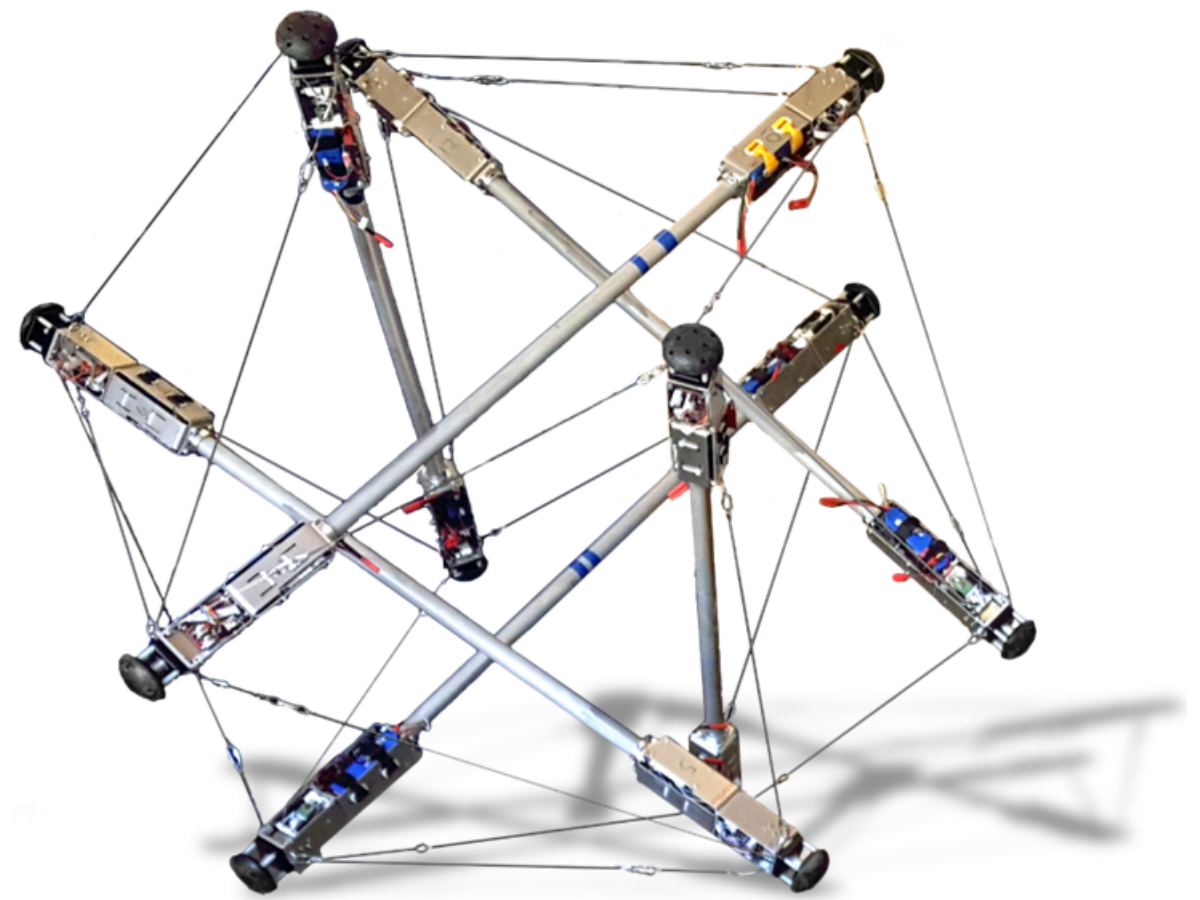}
    \caption{
        \label{fig:superball}
        The \SB{} tensegrity robot. This robot is composed of six identical rods
        and 24 cables, 12 of which can be actuated using the motors connected to
        the ends of each rod. This work uses the onboard IMUs and motor encoders
        on each of the end caps.
    }
\end{figure}

The \SB{} tensegrity robot, shown in Figure~\ref{fig:superball}, is designed to
explore a new class of planetary exploration robot which is able to deploy from
a compact launch volume, land at high speeds without the use of air-bags, and
provide robust surface mobility~\cite{bruce2014design,sabelhaus2015system}.
\SB{} is made up of six identical rods suspended together by 24 cables to form
an icosahedron geometry. The cables have a linear spring attached in-line,
giving the system series elasticity and compliance. Each rod of \SB{} is
comprised of two modular robotic platforms (end caps), as defined
in~\cite{bruce2014design}. Each end cap is equipped with an inertial measurement
unit (IMU), a motor with encoder, and a control board.  Since there is only one
motor per end cap, \SB{} can only actuate 12 of its 24 cables, and this by
definition makes the system underactuated. A basic locomotion strategy for \SB{}
is to contract a cable that is a part of the current ground face of the robot,
thus shrinking the triangle base and causing the robot to tip
over~\cite{bruce2014design}. More sophisticated locomotion may still follow this
strategy, but may also utilize other cables to allow for greater efficiency in
transitioning to the next ground face. These sophisticated strategies are
difficult to hand-engineer, but we demonstrate that they can be learned using
policy search algorithms such as MDGPS.

\subsection{Mirror Descent Guided Policy Search}
\label{sec:mdgps}

\begin{figure}
    \centering
    \setlength{\unitlength}{0.5\columnwidth}
    \begin{picture}(1.0,1.6)
        \put(-0.50,-0.03){\includegraphics[width=\linewidth]{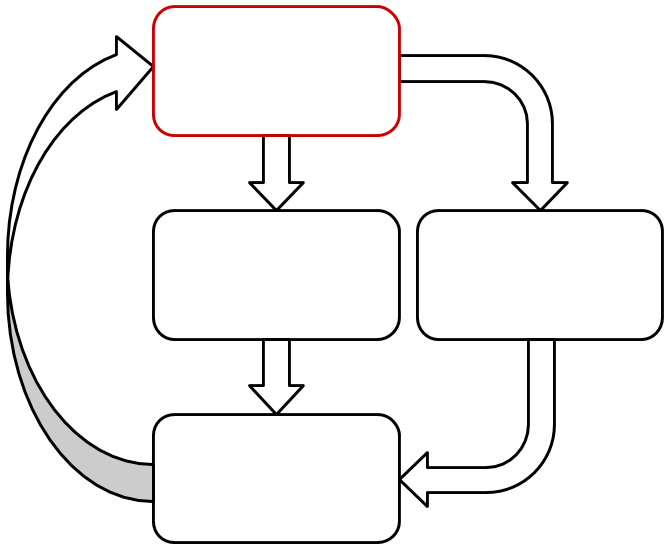}}
        \put(0.00,1.50){Collect}
        \put(0.00,1.42){episodes}
        \put(0.00,1.34){$\{\tau\}$ from}
        \put(0.00,1.26){robot}
        \put(0.31,1.28){\includegraphics[width=0.18\linewidth]{superball.png}}
        \put(0.07,0.87){Fit dynamics}
        \put(0.07,0.78){$p(\mathbf{x}_{t+1}|\mathbf{x}_t,\mathbf{u}_t)$}
        \put(0.07,0.68){to episodes $\{\tau\}$}
        \put(0.78,0.87){Train global policy}
        \put(0.78,0.78){$\pi_\theta(\mathbf{u}_t|\mathbf{o}_t)$}
        \put(0.78,0.68){using $\{\tau\}$}
        \put(1.12,0.66){\includegraphics[width=0.17\linewidth,height=0.03\textheight]{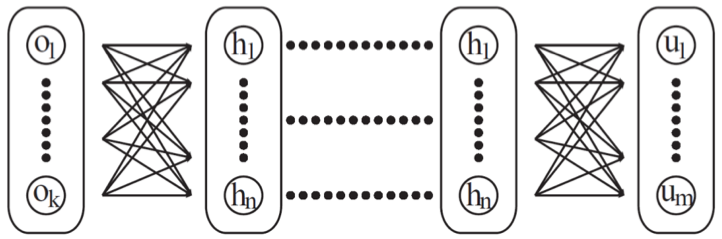}}
        \put(0.00,0.27){Improve $p_i(\mathbf{u}_t|\mathbf{x}_t)$}
        \put(0.00,0.17){using dynamics and}
        \put(0.00,0.07){constrained by $\pi_\theta$}
    \end{picture}
    \caption{
        \label{fig:mdgps}
        Diagram of the MDGPS algorithm described in Algorithm~\ref{alg:mdgps}.
        As described in Section~\ref{sec:chains}, our approach differs from
        standard MDGPS in the highlighted box at the top. The choice of whether
        to collect samples from the local policies $p_i$ or global policy
        $\pi_\theta$ is a hyperparameter.
    }
\end{figure}

Policy search algorithms aim to find a good policy by directly searching through
the space of policy parameters. Formally, we wish to find a setting of the
policy parameters $\params$ to optimize the policy
$\policy_{\params}(\at|\obs_t)$ with respect to the expected cost. In the
finite-horizon episodic setting, the expected cost under the policy is given by
$\return(\params)=\sum_{t=1}^{T}\mathbb{E}_{\policy_{\params}}[\cost(\st,\at)]$,
where $\cost(\st,\at)$ is the cost function. Here $\st$ denotes the state of our
system at time $t$, $\obs_t$ denotes the observation of the state, and $\at$
denotes the action.

Guided policy search algorithms use supervised learning to train the policy,
with supervision coming from several local policies $\trajdist_i(\at|\st)$ that
are optimized to succeed only from a specific initial state of the task, using
full state information. This is much simpler than the goal of the global
policy, which is to succeed under partial observability from any initial
condition sampled from the initial state distribution. These simplifications
allow the use of simple and efficient methods for training the local policies,
such as trajectory optimization methods when there is a known model, or
trajectory-centric reinforcement learning (RL) methods~\cite{la-lnnpg-14}. The
specific algorithm used in this work is MDGPS, which interprets GPS as
approximate mirror descent on $\return(\params)$~\cite{ml-gpsam-16}. The local
policies are optimized to minimize $\return(\params)$ subject to a bound on the
KL-divergence between the local policy $\trajdist_i$ and the global policy
$\bar\policy_{\params{i}}$, following previous
work~\cite{bagnell2003covariant,ps-rlmsp-08,pma-reps-10,slmja-trpo-15}.
Optimizing the global policy is done using the samples collected in the current
iteration, which are used in a supervised fashion to approximately minimize the
divergence between the global policy and the local policies.

\setlength{\textfloatsep}{12pt}
\begin{algorithm}[tb]
    \caption{Mirror descent guided policy search (MDGPS)}
    \label{alg:mdgps}
    \begin{algorithmic}[1]
        \FOR{iteration $k = 1$ to $K$}
            \STATE Run either each $\trajdist_i$ or $\policy_{\params}$ to
            generate episodes $\{\traj\}$
            \STATE Set
            $\trajdist_i\leftarrow\argmin_{\hat\trajdist_i}\mathbb{E}_{\hat\trajdist_i}[\cost(\traj)]~s.t.~D_{KL}(\hat\trajdist_i\|\policy_{\params{i}})\leq\epsilon$
            \STATE Train $\policy_{\params}$ using supervised learning on
            $\{\traj\}$
        \ENDFOR
    \end{algorithmic}
\end{algorithm}

The generic MDGPS algorithm is summarized in Algorithm~\ref{alg:mdgps} and
depicted in Figure~\ref{fig:mdgps}. In the top box, corresponding to line 2,
samples are collected by running either the global policy or the local
policies. This choice between ``on-policy'' and ``off-policy'' sampling is
set by the user, and previous work has noted that ``on-policy'' sampling aids in
the generalization capabilities of the final policy~\cite{ml-gpsam-16}. In the
center and bottom boxes, corresponding to line 3, the algorithm improves the
local policies using an LQR-based update: the samples are first used to fit
time-varying linear dynamics for each local policy, and these fitted dynamics
are then used with a KL-constrained LQR optimization to update the
linear-Gaussian local policies. This corresponds to a simple model-based
trajectory-centric RL method, and further details can be found in prior
work~\cite{la-lnnpg-14}. In the right box and on line 4, the global policy
$\policy_\params(\at|\st)$ is updated using supervised learning, with the
training data corresponding to the observations along the episodes, with actions
given by the updated local policies. This causes the global policy to ``catch
up'' to the local policies, improving its behavior for the next iteration.

\section{Optimizing Periodic Gaits with MDGPS}
\label{sec:chains}

The efficiency and speed of model-based policy optimization methods, such as the
LQR-based method used to optimize the local policies in MDGPS, is due in large
part to the fact that dynamic programming can allow for large changes to a
policy that would be impractical with purely sample-based model-free
methods~\cite{la-lnnpg-14}. However, in complex stochastic domains, such as the
contact-rich dynamical system of a rolling tensegrity robot, the accumulation of
uncertainty and variability under an unstable policy and compounding modeling
errors can make it difficult to apply dynamic programming over the long horizons
needed to establish a periodic gait. Specifically, the LQR-based update we use
fits a time-varying linear-Gaussian dynamics model
$\passivedyn(\state_{t+1}|\st,\at)=\mathcal{N}(\fxt\st+\fut\at,\noise_t)$ and
simulating this model forward becomes increasingly inaccurate with the system
complexity and the length of the horizon.

We demonstrate this effect through our experimental results in
Section~\ref{sec:simresults} and in the appendix. In this section, we describe
how we can obtain a policy with stable periodic behavior by first splitting the
task across multiple policies, each optimized over a smaller time segment where
it is easier to apply dynamic programming, and establishing good behavior across
the range of states visited by these policies. Following the framework of GPS
algorithms, we subsequently learn a global policy that can generalize the
behavior of the local policies and encapsulate a successful periodic gait.

GPS algorithms such as MDGPS use supervised learning to learn a global policy,
where the supervision comes from several local policies $\trajdist_i(\at|\st)$,
$i\in\{1,\ldots,C\}$. Each local policy is trained from a different initial
state, where $C$ is the chosen number of initial states. Each local policy is
optimized over $T^{\trajdist}$ time steps, and we wish to learn a global policy
$\policy_{\params}(\at|\obs_t)$ that can succeed by generalizing the behavior of
these local policies over an episode of length $T^{\policy}$. In manipulation
tasks, we usually set $T^{\policy}=T^{\trajdist}$, depending on the horizon that
we want the task to be accomplished by~\cite{lfda-eetdv-16}. In locomotion
tasks, we ideally want the global policy to exhibit continuous successful
behavior, i.e., $T^{\policy}=\infty$, and we can empirically determine
$T^{\trajdist}$ based on the amount of supervision the global policy needs to
learn a continuous periodic gait. In our work, we simply initialize
$T^{\trajdist}$ to a short horizon, and, should this fail, we restart the
experiment with an increased $T^{\trajdist}$. We repeat this process until the
global policy learns a successful locomotion gait.

If the required $T^{\trajdist}$ is long, as is the case for the \SB{} locomotion
task, it is difficult to optimize a local policy over this time horizon due to
the accumulation of uncertainty and errors described earlier. In our method, we
instead learn $L$ local policies $\trajdist_i^1,\ldots,\trajdist_i^L$ for each
initial state $i$, each optimized for $T^{\trajdist}/L$ time steps. For the
local policies $\trajdist_i^j$, $j\in\{2,\ldots,L\}$, we set the initial state
$\state_0^j$ to be the final state of the preceding local policy, i.e.,
$\state_{T^{\trajdist}/L}^{j-1}$. This amounts to training local policies in a
sequential fashion, where the $L$ local policies together are optimized over
$T^{\trajdist}$ time steps. In practice, we found that the variance of
$\state_{T^{\trajdist}/L}^{j-1}$ is too high at the beginning of training for
$\trajdist_i^j$ to be successful, so we train $\trajdist_i^{j-1}$ until it
stabilizes before starting to train $\trajdist_i^j$.

In previous work, the motivation behind choosing multiple initial states was so
that the global policy could succeed from all of these initial states, and
ideally generalize to other initial states as well~\cite{lfda-eetdv-16}. In our
work, we found that choosing multiple initial states is also beneficial in
learning a stable periodic gait compared to having just one initial state, and
Section~\ref{sec:simresults} demonstrates this difference. The reason for this
is that, to achieve the same amount of supervision using only one initial state,
the sequence of local policies $\trajdist_1^1,\ldots,\trajdist_1^L$ must be much
longer, and training a longer sequence increases the chances of divergence and
instability in the behavior of the local policies, due to compounding variance
in the starting states of later policies, compared to training shorter sequences
from several stable initial states.

\setlength{\textfloatsep}{12pt}
\begin{algorithm}[tb]
    \caption{MDGPS with sequential local policies}
    \label{alg:mdgpsseq}
    \begin{algorithmic}[1]
        \FOR{iteration $k = 1$ to $K$}
            \FOR{$i = 1$ to $C$}
                \STATE $S_i\leftarrow\{\}$
                \FOR{the desired number of episodes}
                    \STATE $x_0\leftarrow$ initial state $i$
                    \FOR{$l = 1$ to $L$}
                        \STATE Run either $\trajdist_i^l$ or $\policy_{\params}$ to
                        generate episode $\traj$
                        \STATE $S_i\leftarrow{S_i}\cup\{\traj\}$
                        \STATE $x_0\leftarrow$ end state of $\traj$
                    \ENDFOR
                \ENDFOR
                \FOR{$l = 1$ to $L$}
                    \STATE $\trajdist_i^l\leftarrow\argmin_{\hat\trajdist_i^l}\mathbb{E}_{\hat\trajdist_i^l}[\cost(\traj)]~s.t.~D_{KL}(\hat\trajdist_i^l\|\policy_{\params{i}})\leq\epsilon$
                \ENDFOR
            \ENDFOR
            \STATE Train $\policy_{\params}$ using supervised learning on
            $\bigcup_iS_i$
        \ENDFOR
    \end{algorithmic}
\end{algorithm}

Our method is detailed in Algorithm~\ref{alg:mdgpsseq}. On line 7, we collect
samples from either the local policies or global policy. In our work, we start
by sampling from the local policies, and we switch to sampling from the global
policy after a fixed number of iterations, which we set empirically based on the
performance and stability of the local policies. This allows for the local
policies to establish good rolling behavior at the beginning, and then for the
global policy to generalize this into continuous rolling. On line 8, we store
the collected episode, and in order to run the policies sequentially, we set the
starting state of the next policy to be the end state of the episode $\traj$
collected from the current policy. The rest of the algorithm is identical to
standard MDGPS -- note that lines 3 through 11, 12 through 14, and 16 in
Algorithm~\ref{alg:mdgpsseq} correspond to line 2, 3, and 4, respectively, in
Algorithm~\ref{alg:mdgps}.

\section{Learning Locomotion for \SB{}}
\label{sec:superball}

For \SB{} locomotion, we chose six initial states that correspond to the robot
resting on each of its stable ground faces. We set $T^{\trajdist}=100$ and
$L=2$, so from each initial state $i$, two local policies $\trajdist_i^1$ and
$\trajdist_i^2$ are optimized over 50 time steps each, where each time step is
\SI{0.1}{\second}. We found that a fully optimized local policy can perform
about two transitions from one ground face to the next face over
\SI{5}{\second}. We also found it helpful in our work to not begin training
$\trajdist_i^2$ until $\trajdist_i^1$ trains for several iterations, so that
$\state_{50}^1$, which is $\state_0^2$, is more stable and has lower variance.
We show in the appendix that training two local policies in this fashion is much
more sample efficient than training one local policy for \SI{10}{\second}, and
produces smoother and faster rolling behavior. In total, in our work, 12 local
policies are trained in sequences of two, and this provided the global policy
the supervision required to learn to roll continuously.

\subsection{Kinematic Constraints for Safe Actions}
\label{sec:superball_constraints}

One of the challenges with automated policy learning is that all requirements
for the policy are generally encoded in the cost function, including task-level
objectives such as desired rolling direction and hard constraints such as
safety. Due to the structure of tensegrity systems, unsafe actuation of the
motors can place the robot into configurations with unacceptable risk of cable
or motor failure. The configurations associated with such high-tension
conditions are difficult to encode analytically and even more difficult to
balance against primary task objectives in the cost function. We therefore adopt
a simple safety constraint approach to enable safe learning and policy execution
on the \SB{} hardware. This approach is challenging to use with hand-engineered
policies, which often exploit unsafe but effective actions to quickly yield a
locomotion gait. However, the approach is much easier to adopt with learned
policies, as it naturally embeds itself into the training procedure.

Specifically, we estimate the cable tensions for a particular set of actuator
positions using a simple forward kinematics model of \SB{}.  We repeat this
process for many different randomly generated motor settings and store all sets
of motor positions for which all cable tensions of the robot are below the
acceptable threshold. Then, when the policy outputs an action, we use the Fast
Library for Approximate Nearest Neighbors (FLANN)~\cite{flannsoftware} to compute
and command the nearest ($\cost_1$ norm) safe action. Computing the cable tensions
for a given set of motor positions takes a few milliseconds using forward
kinematics. As this is easily parallelized, we constructed a database containing
about 100 million motor positions deemed safe in a few hours. At runtime,
finding the nearest neighbor action takes roughly \SI{200}{\micro\second} and is
easily embedded into both training and testing without disrupting the command
frequency of \SI{10}{\hertz}.

By separating this safety constraint from the cost function, we avoid the need
to tune the parameters of the cost function to weigh the opposing objectives of
speed and hardware safety. Furthermore, we encode safety as a hard constraint
using this method, and we directly prevent unsafe actions rather than just
penalizing them. Utilizing these kinematic constraints is extremely helpful in
transferring policies learned in simulation directly onto the real robot. In
simulation, the physical limits are less restrictive, and going beyond the safe
limits does not have any adverse effects, so it is possible to train a policy
that exploits these inaccuracies in order to succeed. We can make sure this does
not happen by enforcing stronger constraints on what actions the policy can
output, and it is more likely that the policy trained with these constraints in
simulation will not fail on the real robot, or even worse, cause damage and
hardware problems.

\section{Experimental Results}
\label{sec:results}

In our experiments, we aim to answer several questions. First, can we learn
policies in simulation that allow the simulated \SB{} to roll efficiently under
various settings of terrain, gravity, sensor noise, and robot parameters?
Second, do our learned feedback policies generalize better to changes in
environmental and robot parameters compared to open-loop learned policies and
hand-engineered controllers? Finally, can we transfer the learned policy from
simulation to the real robot, and have the real robot roll?

Section~\ref{sec:setup} details our experimental setup. In
Section~\ref{sec:simresults}, we test in simulation the efficiency of the
learned policies and make comparisons to establish the importance of learning
and feedback. In Section~\ref{sec:realresults}, we evaluate a learned policy on
the real robot. The appendix includes an additional experiment that illustrates
the benefit of learning sequential policies, each over shorter time horizons.

\subsection{Experimental Setup}
\label{sec:setup}

We encode the state of the system $\st$ as the position and velocity of each of
the 12 bar endpoints of \SB{}, and the position and velocity of each of the 12
motors, measured in radians, for a total dimensionality of 96. We experimented
with two different representations for the observation $\obs_t$. The ``full''
36-dimensional observation includes motor positions, and also uses elevation and
rotation angles and angular velocities calculated from the robot's
accelerometers and magnetometers. The ``limited'' observation is 12-dimensional
and only uses the acceleration measurement along the bar axis from each of the
accelerometers. We found that interfering magnetic fields near the testing
grounds at NASA Ames cause the magnetometers to be unreliable and difficult to
calibrate, and because of this, the policy using the limited observation is much
easier to transfer on to the real robot. The action $\at$ is the instantaneous
desired position of each motor, which may not be reached if the position is far
away.

Each local policy required about 200 episodes before it was reliably successful.
Simultaneously during training of the local policies, we learn a global policy,
which in our work is a deep neural network with three hidden layers of 64
rectified linear units (ReLU) each, using the same samples. Our cost function
$l(\st,\at)$ is simply the negative average velocity of the bar endpoints of the
robot, which corresponds to the linear velocity of the center of mass. The local
policies are therefore trained to roll as quickly as possible.

For completeness, the entire learning process took approximately \SI{2}{\hour}
on a CUDA-enabled quad-core Intel i7 computer utilizing \SI{4}{\gibi\byte} of
system memory. This is with each simulation set running at faster than real time
and with the kinematic constraints mentioned in
Section~\ref{sec:superball_constraints}. Without CUDA, the learning process
takes approximately 20\% longer.

\subsection{Results in Simulation}
\label{sec:simresults}

\begin{table*}[t]
    \centering
    \resizebox{\textwidth}{!}{%
        \begin{tabular}{|llrrrrr|}
            \hline
            \multicolumn{2}{|c}{}                                                                                                     & \multicolumn{3}{|c|}{}                                                                                                                                                                                                    & \multicolumn{2}{c|}{} \\
            \multicolumn{2}{|c}{}                                                                                                     & \multicolumn{3}{|c|}{\multirow{-2}{*}{\textbf{Our Method}}}                                                                                                                                                               & \multicolumn{2}{c|}{\multirow{-2}{*}{\textbf{Open-Loop}}} \\
            \cline{3-7}
            \multicolumn{2}{|c}{}                                                                                                     & \multicolumn{1}{|r|}{Full Observation,}                                 & \multicolumn{1}{r|}{Limited Observation,}                              & \multicolumn{1}{r|}{Limited Observation,}                              & \multicolumn{1}{r|}{Mean Actions from}                           & Hand-Engineered \\
            \multicolumn{2}{|c}{}                                                                                                     & \multicolumn{1}{|r|}{Six Initial States}                                & \multicolumn{1}{r|}{Six Initial States}                                & \multicolumn{1}{r|}{One Initial State}                                 & \multicolumn{1}{r|}{Best Learned Policy}                         & Punctuated Rolling \\
            \hline
            \multicolumn{2}{|l}{}                                                                                                     & \multicolumn{1}{|r|}{}                                                  & \multicolumn{1}{r|}{}                                                  & \multicolumn{1}{r|}{}                                                  & \multicolumn{1}{r|}{}                                            & \\
            \multicolumn{2}{|l}{\multirow{-2}{*}{\textbf{Normal Conditions}}}                                                         & \multicolumn{1}{|r|}{\multirow{-2}{*}{$\mathbf{25.307\pm0.309}$}}       & \multicolumn{1}{r|}{\multirow{-2}{*}{$24.141\pm0.352$}}                & \multicolumn{1}{r|}{\multirow{-2}{*}{$20.008\pm0.871$}}                & \multicolumn{1}{r|}{\multirow{-2}{*}{$\mathit{25.076\pm0.078}$}} & \multirow{-2}{*}{$10.266\pm0.071$} \\
            \hline
            \rowcolor[HTML]{D3D3D3}
            \cellcolor[HTML]{D3D3D3}                                   & \multicolumn{1}{|r}{\cellcolor[HTML]{D3D3D3}Hilly}           & \multicolumn{1}{|r|}{\cellcolor[HTML]{D3D3D3}$\mathit{6.025\pm2.835}$}  & \multicolumn{1}{r|}{\cellcolor[HTML]{D3D3D3}$\mathbf{9.568\pm5.197}$}  & \multicolumn{1}{r|}{\cellcolor[HTML]{D3D3D3}$3.124\pm1.083$}           & \multicolumn{1}{r|}{\cellcolor[HTML]{D3D3D3}$3.069\pm2.201$}     & $1.734\pm0.411$ \\
            \rowcolor[HTML]{D3D3D3}
            \cellcolor[HTML]{D3D3D3}                                   & \multicolumn{1}{|r}{\cellcolor[HTML]{D3D3D3}Uphill}          & \multicolumn{1}{|r|}{\cellcolor[HTML]{D3D3D3}$\mathbf{18.547\pm0.231}$} & \multicolumn{1}{r|}{\cellcolor[HTML]{D3D3D3}$\mathit{16.107\pm0.809}$} & \multicolumn{1}{r|}{\cellcolor[HTML]{D3D3D3}$13.573\pm0.174$}          & \multicolumn{1}{r|}{\cellcolor[HTML]{D3D3D3}$7.721\pm0.236$}     & $8.136\pm0.026$ \\
            \rowcolor[HTML]{D3D3D3}
            \multirow{-3}{*}{\cellcolor[HTML]{D3D3D3}\textbf{Terrain}} & \multicolumn{1}{|r}{\cellcolor[HTML]{D3D3D3}Downhill}        & \multicolumn{1}{|r|}{\cellcolor[HTML]{D3D3D3}$\mathbf{32.896\pm0.275}$} & \multicolumn{1}{r|}{\cellcolor[HTML]{D3D3D3}$\mathit{29.970\pm0.858}$} & \multicolumn{1}{r|}{\cellcolor[HTML]{D3D3D3}$21.963\pm2.403$}          & \multicolumn{1}{r|}{\cellcolor[HTML]{D3D3D3}$27.661\pm0.136$}    & $11.264\pm0.091$ \\
            \hline
                                                                       & \multicolumn{1}{|r}{10\%}                                    & \multicolumn{1}{|r|}{$\mathbf{19.505\pm0.746}$}                         & \multicolumn{1}{r|}{$0.804\pm0.103$}                                   & \multicolumn{1}{r|}{$0.776\pm0.025$}                                   & \multicolumn{1}{r|}{$\mathit{18.024\pm2.356}$}                   & $11.044\pm0.054$ \\
                                                                       & \multicolumn{1}{|r}{50\%}                                    & \multicolumn{1}{|r|}{$\mathbf{23.331\pm0.871}$}                         & \multicolumn{1}{r|}{$9.671\pm0.550$}                                   & \multicolumn{1}{r|}{$0.996\pm0.007$}                                   & \multicolumn{1}{r|}{$\mathit{19.673\pm3.244}$}                   & $10.310\pm0.010$ \\
            \multirow{-3}{*}{\textbf{Gravity}}                         & \multicolumn{1}{|r}{200\%}                                   & \multicolumn{1}{|r|}{$\mathit{27.600\pm2.307}$}                         & \multicolumn{1}{r|}{$\mathbf{30.095\pm1.055}$}                         & \multicolumn{1}{r|}{$16.422\pm4.068$}                                  & \multicolumn{1}{r|}{$24.865\pm0.190$}                            & $9.845\pm0.009$ \\
            \hline
            \rowcolor[HTML]{D3D3D3}
            \cellcolor[HTML]{D3D3D3}                                   & \multicolumn{1}{|r}{\cellcolor[HTML]{D3D3D3}Heavy}           & \multicolumn{1}{|r|}{\cellcolor[HTML]{D3D3D3}$12.521\pm1.710$}          & \multicolumn{1}{r|}{\cellcolor[HTML]{D3D3D3}$\mathbf{14.561\pm0.079}$} & \multicolumn{1}{r|}{\cellcolor[HTML]{D3D3D3}$\mathit{12.972\pm0.110}$} & \multicolumn{1}{r|}{\cellcolor[HTML]{D3D3D3}$1.081\pm0.019$}     & $10.550\pm0.003$ \\
            \rowcolor[HTML]{D3D3D3}
            \multirow{-2}{*}{\cellcolor[HTML]{D3D3D3}\textbf{Robot}}   & \multicolumn{1}{|r}{\cellcolor[HTML]{D3D3D3}End Cap Failure} & \multicolumn{1}{|r|}{\cellcolor[HTML]{D3D3D3}$\mathbf{23.215\pm2.927}$} & \multicolumn{1}{r|}{\cellcolor[HTML]{D3D3D3}$\mathit{21.595\pm0.253}$} & \multicolumn{1}{r|}{\cellcolor[HTML]{D3D3D3}$19.144\pm0.207$}          & \multicolumn{1}{r|}{\cellcolor[HTML]{D3D3D3}$17.901\pm1.967$}    & $10.222\pm0.013$ \\
            \hline
                                                                       & \multicolumn{1}{|r}{0\%}                                     & \multicolumn{1}{|r|}{$\mathbf{27.250\pm0.203}$}                         & \multicolumn{1}{r|}{$\mathit{25.798\pm0.200}$}                         & \multicolumn{1}{r|}{$20.918\pm0.213$}                                  & \multicolumn{1}{r|}{}                                            & \\
            \multirow{-2}{*}{\textbf{Added Noise}}                     & \multicolumn{1}{|r}{20\%}                                    & \multicolumn{1}{|r|}{$8.739\pm7.031$}                          & \multicolumn{1}{r|}{$\mathbf{18.095\pm0.406}$}                         & \multicolumn{1}{r|}{$\mathit{17.613\pm0.400}$}                                  & \multicolumn{1}{r|}{\multirow{-2}{*}{N/A}}                       & \multirow{-2}{*}{N/A} \\
            \hline
        \end{tabular}%
    }
    \caption{
        \label{table:distance}
        Average distances in meters traveled in simulation using the policies
        learned through our method with varying observation representations and
        local policy training schemes, the open-loop mean actions from the
        learned policy that performs best under training conditions, and the
        hand-engineered open-loop policy. Results are averaged across five
        trials of one minute each for a variety of terrain, gravity, noise, and
        robot settings. ``Normal Conditions'' are the training conditions, which
        are flat terrain, 100\% gravity, 10\% added noise to the input, and
        normal robot parameters. For the terrain settings, the hilly terrain
        contained hills that were \SI{0.1}{\meter} tall, and the uphill and
        downhill terrains had slopes of \SI{0.1}{\radian}. When varying one
        setting, all other settings remain the same as during training time. We
        do not test the open-loop controllers with varying input noise, because
        these controllers do not have any input. Bolded numbers indicate the
        farthest distance traveled for any given condition, and italicized
        numbers are the second farthest. Note that the first two learned
        policies generally outperform all other controllers, demonstrating the
        benefits of our method combined with multiple initial states.
    }
\end{table*}

The \SB{} simulation that we use is built off of the NTRT open-source project.
Aside from speed and efficiency benefits, the simulation also allows us to
systematically and easily vary the parameters of the robot and environment.

Our results for testing the learned policies against a range of environmental
and robot parameters are presented in Table~\ref{table:distance}. We report the
average distance traveled over five trials of \SI{60}{\second} for three
policies learned with MDGPS, an open-loop policy that outputs the mean actions
from the learned policy that performs best under training conditions, and an
open-loop hand-engineered controller. The policies learned with MDGPS use either
the 36-dimensional observation or the 12-dimensional observation. To demonstrate
the benefit of multiple initial states, we also learn a policy using the
accelerometer observation and a long sequence of local policies, still trained
over \SI{5}{\second} each, from one initial state. The open-loop hand-engineered
controller is designed to follow the basic locomotion strategy described in
Section~\ref{sec:tensegrity}.

We systematically vary a suite of settings in simulation: the ground terrain,
which can be flat, uneven, or sloped up or downhill; the strength of the
gravitational field, which can be 10\%, 50\%, 100\%, or 200\% of Earth's
gravity; the noise level on the inputs to the policy, which can be 0\%, 10\%,
and 20\%; and various parameters of the robot. The noise we
incorporate into the robot's sensor readings includes Gaussian noise, with
variance for each sensor type proportional to the range of the sensor, and
randomly dropped sensor readings. Details about this noise can be found in the
appendix. For robot parameters, we increase the mass of the rods and decrease
the maximum motor velocities to create a heavier robot, and we simulate the
failure of a specific end cap by dropping all sensors readings and motor
commands.

The learned policy with full observation performs best under training
conditions, and also demonstrates the best generalization across terrain and
gravity settings as well as the noiseless condition. The learned policy with
limited observation, though slightly slower than the policy with full
observation, adapts better to the heavier robot as well as to motor failure, and
also performs significantly better under heavy noise. These conditions are
important for successful transfer to the real world setting, since the
parameters of the physical \SB{} robot naturally differ from the simulation, and
the imperfections in the hardware result in sensor and motor unreliability. The
learned policy from the single long sequence of local policies does not perform
as well as the other learned policies, because it is prone to diverge and become
unstable, at which point it fails. In training this policy, we were unable to
successfully train the fifth local policy to continue the gait, due to both the
build-up in variance in the starting states of the local policies and the
divergence in the behavior of the local policies from the desired periodic gait.
The global policy learns a gait from the first four local policies, but the
resulting behavior is less efficient and also significantly less stable.

The open-loop mean actions from the learned policy demonstrate comparable
results under the training conditions, as expected, but its performance falls
off drastically for most of the other conditions. Because a controller for \SB{}
would ideally work under different terrains and in the presence of hardware
issues, this shows that it is imperative that we use a closed-loop
representation to ensure successful and reliable locomotion. The hand-engineered
open-loop controller rolls consistently across most conditions, but is
significantly slower than the three other policies. This emphasizes the benefits
of learning and the difficulty in hand-engineering good controllers, as this
controller was carefully designed but still sacrifices speed and other important
benefits, such as robot safety, for reliability. Most notably, this
hand-engineered controller has caused cable and motor failure on the physical
\SB{} robot in the past, which is why we did not compare against it for the real
robot results.

In summary, these results show that all learned policies substantially
outperform the hand-engineered rolling controller, and the closed-loop neural
network policies outperform the open-loop baselines in almost all conditions,
indicating the benefits of both learning and feedback in \SB{} locomotion. Our
method is able to learn successful and efficient policies even with the limited
sensory observations provided by only \SB{}'s accelerometers, and the learned
policies demonstrate generalization to unseen conditions representative of what
a planetary exploration rover might encounter, such as changing terrains,
unstable levels of noise, and hardware failure. The addition of input noise
during training encourages this generalization, and results in learned policies
with similar levels of reliability as the hand-engineered controller, though
significantly faster and less likely to cause hardware failure on the real
robot.

\subsection{Results in the Real World}
\label{sec:realresults}

The limited observation policy that was trained in simulation was successfully
run directly on the real \SB{} with no tuning or changes. We then compared how
this policy performed against an open-loop policy that only outputs motor
actions with no feedback. These motor actions were derived as the average time
sampled actions from the policy running in simulation under training conditions.
Both policies were run on the physical \SB{} robot on flat terrain. Videos of
the training process and experiments can be found on
\url{http://rll.berkeley.edu/drl\_tensegrity}.

\begin{figure}
    \centering
    \setlength{\unitlength}{0.5\columnwidth}
    \includegraphics[width=\linewidth]{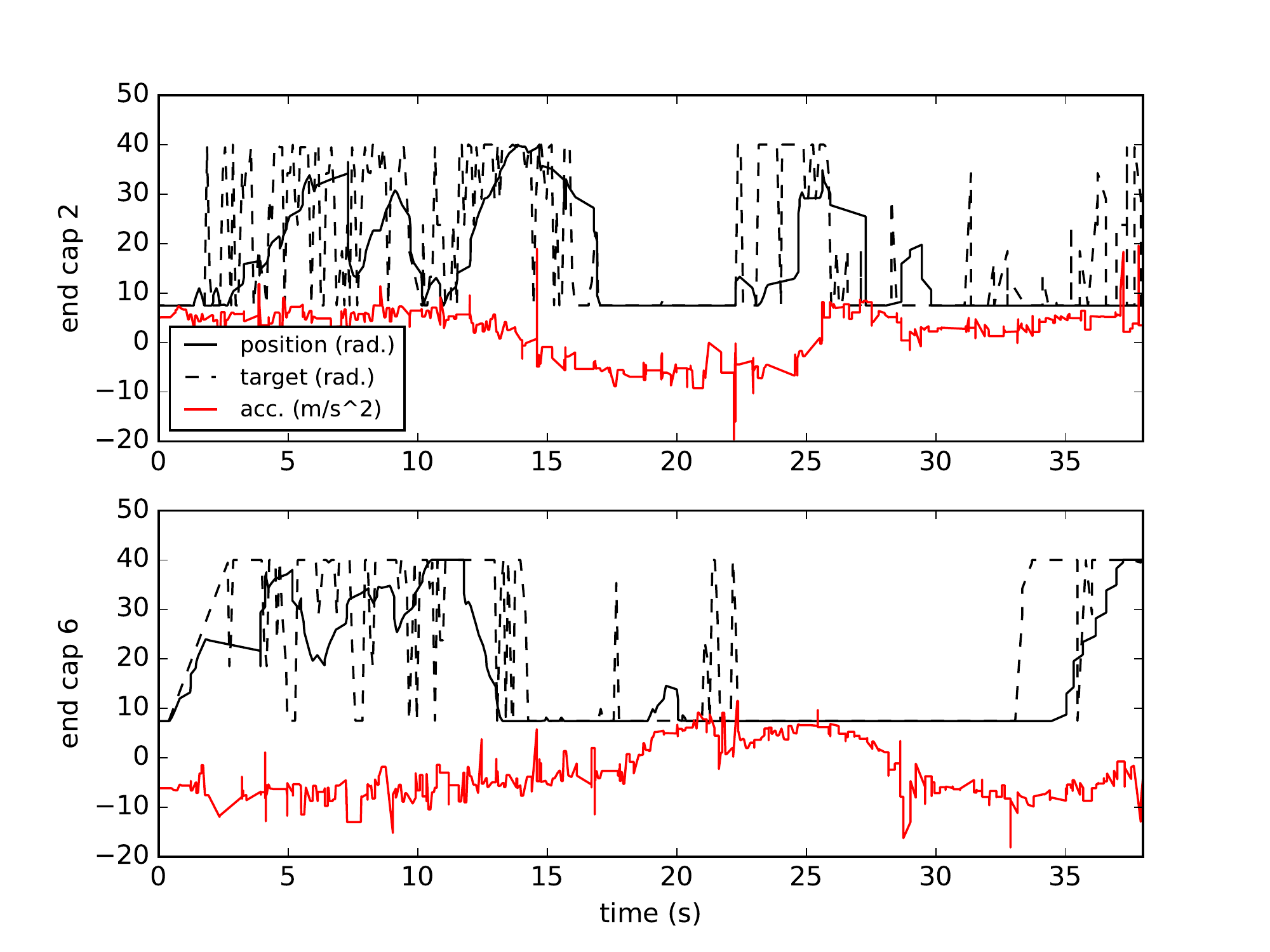}
    \caption{
        \label{fig:commands}
        This plot shows actual motor positions (solid top), commanded target
        motor positions (black dotted), and single axis accelerometer data
        (solid bottom) for two disconnected end caps over the first
        \SI{40}{\second} of a trial. The learned policy commands target motor
        positions using accelerometer feedback. The actual motor positions lag
        behind due to motor dynamics and network UDP packet loss.
    }
\end{figure}

Over three trials of \SI{100}{\second} each, using the learned policy, \SB{}
rolled approximately \SI{12}{\meter}, \SI{9}{\meter}, and \SI{8}{\meter} measured
as the linear distance from the robot's start to final position.
There was a recorded small left turn bias in the robot's over all trajectory during
each trial, most likely due to inconsistent pre-tensioning of individual cables.
\SI{12}{\meter} is around the maximum distance allowed during each trial due to
our limited network range. Despite the differences between the simulated and
physical robot, the policy was able to successfully produce a gait on \SB{} that
is more reliable, and less risky for the hardware, than any previous locomotion
controller for this robot. The learned policy was able to adapt to the physical
\SB{} robot by using feedback from the accelerometers, as seen in
Figure~\ref{fig:commands}.

The open-loop policy was not able to produce any reasonable behavior on the real
robot. The lack of performance exhibited by this policy is due to a mismatch
between the dynamics of the simulated robot and the real robot. Specifically,
the real \SB{} robot has friction in the cable routing system, which introduces
noticeable hysteresis and non-linear behavior into the spring forces. Because of
this, the real robot's individual motor dynamics are not constant nor consistent
for all cables and commanded positions. The policy uses these discrepancies to
achieve faster locomotion in simulation, and the open-loop policy attempts to
mimic this on the real robot, which does not work since the robot cannot reach
the same motor positions over a fixed number of time steps as in simulation. The
learned policy is able to adapt to the physical \SB{} robot by using feedback
from the accelerometers, and it adopts a strategy that applies motor commands
for a longer period of time in order to move the robot into the desired
configuration, thus it is still able to produce a successful locomotion gait.

\section{Discussion and Future Work}

We presented a method for learning locomotion policies for tensegrity robots,
by introducing several improvements to MDGPS that adapt the algorithm to tasks
that require periodic behavior. We demonstrated learned policies for the \SB{}
tensegrity robot in simulation that are efficient and perform well under a
variety of environmental and robot parameters. We also demonstrated a learned
policy that can be transferred directly to the real system to allow the physical
\SB{} robot to roll. We showed that our learned locomotion policies are more
effective and generalize better than open-loop policies and hand-engineered
controllers.

One direction for future work is to extend our method to other locomotion
systems, including other tensegrity robots and legged systems such as
bipeds and quadrupeds. Since our method is specifically designed to tackle the
growth in variance over long-horizon periodic motions, it would be well suited
for learning periodic feedback policies for a wide range of locomotion
platforms. Many of these platforms introduce new challenges such as instability,
which should provide new insights and improvements for our method.

Finally, since learning the locomotion policies in this paper requires only a
few hundred trials, a promising direction for future work is to perform the
learning process itself directly on the physical hardware, which should require
only a few hours of continuous operation. While we chose to use simulated
training in this work, more complex hardware platforms or more elaborate
environments may be difficult to simulate accurately. Furthermore, gradual
changes to the hardware due to damage or wear-and-tear might require retraining
of the policy in situ. Evaluating this application of our method would be an
exciting direction for future work.

{\small
    \paragraph*{Acknowledgments}
    This work was supported in part by the DARPA SIMPLEX program and an NSF
    CAREER award. We appreciate the support, ideas, and feedback from members of
    the Berkeley Artificial Intelligence Research Lab and the Dynamic Tensegrity
    Robotics Lab. We are also grateful to Terry Fong and the NASA Ames
    Intelligent Robotics Group.  
}

\bibliographystyle{IEEEtran}
\bibliography{references,jonathan}
\balance
\newpage

\appendix
\subsection{Additional Experimental Details}

Because we train local policies with full state information $\st$ but learn a
global policy that only receives an observation of the state $\obs_t$ as input,
we are able to train a global policy that operates under partial observability
at test time while maintaining the simplicity of training the local policies on
the full state. This separation between the local policies and global policy
reflects prior work on tasks involving partial observability, where the
intuition is that the local policies are trained in a controlled environment but
the global policy must be able to adapt to a more general
setting~\cite{lfda-eetdv-16}.

In our case, the full state $\st$ can only be obtained through either simulation
or the use of an external state estimator system on the physical \SB{}
robot~\cite{caluwaerts2016esitmation}. In contrast, we choose an observation
$\obs_t$ that can be calculated directly from the sensors on the robot itself.
This greatly simplifies the transfer from simulation to the real robot, as the
learned policy is less prone to overfit to the simulation and takes actions
directly based on the sensor measurements from the physical robot. Furthermore,
because the goal of \SB{} and many other robots is deployment to
unfamiliar, remote environments, the choice of an observation that relies only
on the robot's onboard sensors is very important, as it is unrealistic to expect
the level of information and reliability that an external state estimator can
provide. For a description of the state and observation, as well as the details
and dimensionalities of the sensors, see Section~\ref{sec:setup}.

Because the real-world sensors and actuators are noisy and imperfect, we attempt
to model this in simulation by introducing noise on the input to the policy
during training. We model measurement errors and sensor inaccuracies by adding
Gaussian noise with mean 0 and variance equal to 10\% of the range of the
observation. To model sensor failure, latency, and network issues such as
connection errors, we randomly drop observations 10\% of the time. When the
current observation is dropped, the previous observation is used as the input to
the policy. We found that adding noise improves the generalization capabilities
of the learned policy across conditions such as terrain, gravity, and motor
failure, and we test against these conditions in Section~\ref{sec:simresults}.

\subsection{Additional Experimental Results}

\begin{table}[ht!]
    \centering
    \resizebox{\columnwidth}{!}{%
        \begin{tabular}{|lrr|}
            \hline
                                                   & \multicolumn{1}{|r|}{Two Local Policies,}                              & One Local Policy \\
                                                   & \multicolumn{1}{|r|}{Sequential}                                       & Full \SI{10}{\second} \\
            \hline
            \rowcolor[HTML]{D3D3D3}
            \cellcolor[HTML]{D3D3D3}Episodes Until & \multicolumn{1}{|r|}{\cellcolor[HTML]{D3D3D3}}                         & \cellcolor[HTML]{D3D3D3} \\
            \rowcolor[HTML]{D3D3D3}
            \cellcolor[HTML]{D3D3D3}Convergence    & \multicolumn{1}{|r|}{\multirow{-2}{*}{\cellcolor[HTML]{D3D3D3}$<700$}} & \multirow{-2}{*}{\cellcolor[HTML]{D3D3D3}$<800$} \\
            \hline
            Average Distance                       & \multicolumn{1}{|r|}{}                                                 & \\
            Traveled (m)                           & \multicolumn{1}{|r|}{\multirow{-2}{*}{$3.156$}} & \multirow{-2}{*}{$1.228$} \\
            \hline
        \end{tabular}
    }
    \caption{
        \label{table:chains}
        Comparison of sample efficiency and distance traveled at convergence of
        two local policies trained for \SI{5}{\second} each and one local policy
        trained for the full \SI{10}{\second}. Note that our method of training
        sequential local policies requires fewer samples for convergence and
        demonstrates significantly better performance.
    }
\end{table}

To show that our method of training sequential local policies is effective, we
compared the results of training two sequential local policies for
\SI{5}{\second} each against training one local policy for the full
\SI{10}{\second}, both using the trajectory-centric RL method detailed
in~\cite{la-lnnpg-14}. We record the average distance traveled over five trials
in Table~\ref{table:chains}. The two local policies trained sequentially not
only converges in fewer samples, but also settles into a much more effective
rolling strategy that travels significantly farther over the course of
\SI{10}{\second}. These results show that, by training sequences of local
policies over shorter horizons, we can achieve more efficient locomotion with
fewer samples by decreasing the compounding effects of modeling errors and
unstable policies over time.

\end{document}

%% file: defs.tex
\long\def\ignorethis#1{}

\newcommand{\etal}{{et~al.}\ }

\DeclareMathOperator*{\argmin}{arg\,min}

\newcommand{\trajdist}{p}
\newcommand{\policy}{\pi}

\newcommand{\return}{J}
\newcommand{\params}{\theta}

\newcommand{\cost}{\ell}
\newcommand{\state}{\mathbf{x}}
\newcommand{\obs}{\mathbf{o}}
\newcommand{\action}{\mathbf{u}}

\newcommand{\traj}{\tau}

\newcommand{\second}{\text{s}}

\newcommand{\meter}{\text{m}}

\newcommand{\fxt}{f_{\state t}}
\newcommand{\fut}{f_{\action t}}

\newcommand{\passivedyn}{p}

\newcommand{\noise}{\mathbf{F}}

\newcommand{\st}{\state_t}

\newcommand{\at}{\action_t}